\documentclass{bmvc2k}
\usepackage{multirow}
\usepackage{booktabs}
\usepackage{floatrow}
\newcommand{\sectionvspace}{\vspace{-2em}}
\newcommand{\subsectionvspace}{\vspace{-1.3em}}
\newcommand{\subsectionbelowvspace}{\vspace{-0.5em}}
\newcommand{\subsubsectionvspace}{\vspace{-1em}}
\newfloatcommand{capbtabbox}{table}[][\FBwidth]
\usepackage[normalem]{ulem} 

\title{Cooperative Dual Attention for Audio-Visual Speech Enhancement with Facial Cues} %

\addauthor{Feixiang Wang}{wangfeixiang19@mails.ucas.ac.cn}{1,2}
\addauthor{Shuang Yang}{shuang.yang@ict.ac.c n}{1,2}
\addauthor{Shiguang Shan}{sgshan@ict.ac.cn}{1,2}
\addauthor{Xilin Chen}{xlchen@ict.ac.cn}{1,2}

\addinstitution{
Key Laboratory of Intelligent Information Processing of Chinese Academy of Sciences (CAS), \\
Institute of Computing Technology, CAS, \\
Beijing 100190, China
}

\addinstitution{
University of Chinese Academy of Sciences, \\
Beijing 100049, China
}

\runninghead{Wang., Yang., Shan., Chen.}{DualAVSE}

\begin{document}

\maketitle
\vspace{-1.3em}
\begin{abstract}
In this work, we focus on taking advantage of the facial cues, beyond the lip region, for robust Audio-Visual Speech Enhancement (AVSE). The facial region covers the lip region and furthermore reflects more speech-related attributes obviously, such as gender, skin color, nationality, and so on, which are beneficial for AVSE. 
However, besides the speech-related attributes, there also exist static and dynamic speech-unrelated attributes which always cause speech-unrelated appearance changes during the speaking process.
To address these challenges, we propose a dual attention cooperative framework, named DualAVSE, to ignore speech-unrelated information and fully capture speech-related information with facial cues, then dynamically integrate such information with the audio signal for AVSE. 
Specifically, to capture and enhance the visual speech information beyond the lip region, we propose a spatial attention-based visual encoder to introduce the global facial context and automatically ignore speech-unrelated information for robust visual feature extraction.
Secondly, we introduce a dynamic visual feature fusion strategy by incorporating a temporal-dimensional self-attention module, which enables the model to robustly handle facial variations in the process.
Thirdly, the acoustic noise in the speaking process is always not a stable constant noise, which makes the audio quality in the contaminated speech signal vary in the process. Accordingly, we introduce a dynamic fusion strategy for both the audio feature and visual feature to address this issue. 
By integrating the cooperative dual attention reflected in both the visual encoder and the audio-visual fusion strategy, our model can effectively extract beneficial speech information from both audio and visual cues for AVSE.
We performed a thorough analysis and comparison on different datasets with several settings, including the normal case and hard case when visual information is unreliable or even absent. 
These results consistently show that our model outperforms existing methods under multiple metrics.

\end{abstract}
\sectionvspace
\section{Introduction}
\vspace{-1em}
\label{sec:intro}
Speech enhancement aims to improve the quality and intelligibility of audio speech by suppressing or eliminating background noise in the original noisy speech signals.
It plays a key role for several downstream applications, such as automatic speech recognition~\cite{li2015robust,li2013investigation}, speaker recognition~\cite{el2007evaluation,ortega1996overview}, hearing aids~\cite{venema2006compression,levit2001noise,chern2017smartphone}, and so on.

%
%
Inspired by the McGurk effect~\cite{mcgurk1976hearing} that visual cues play an important role in speech processing in human brains, researchers have begun introducing visual cues to combine with audio for speech enhancement in recent years.
\begin{figure}[htbp]
    \vspace{-1em}
    \centering
    \includegraphics[width=12.5cm]{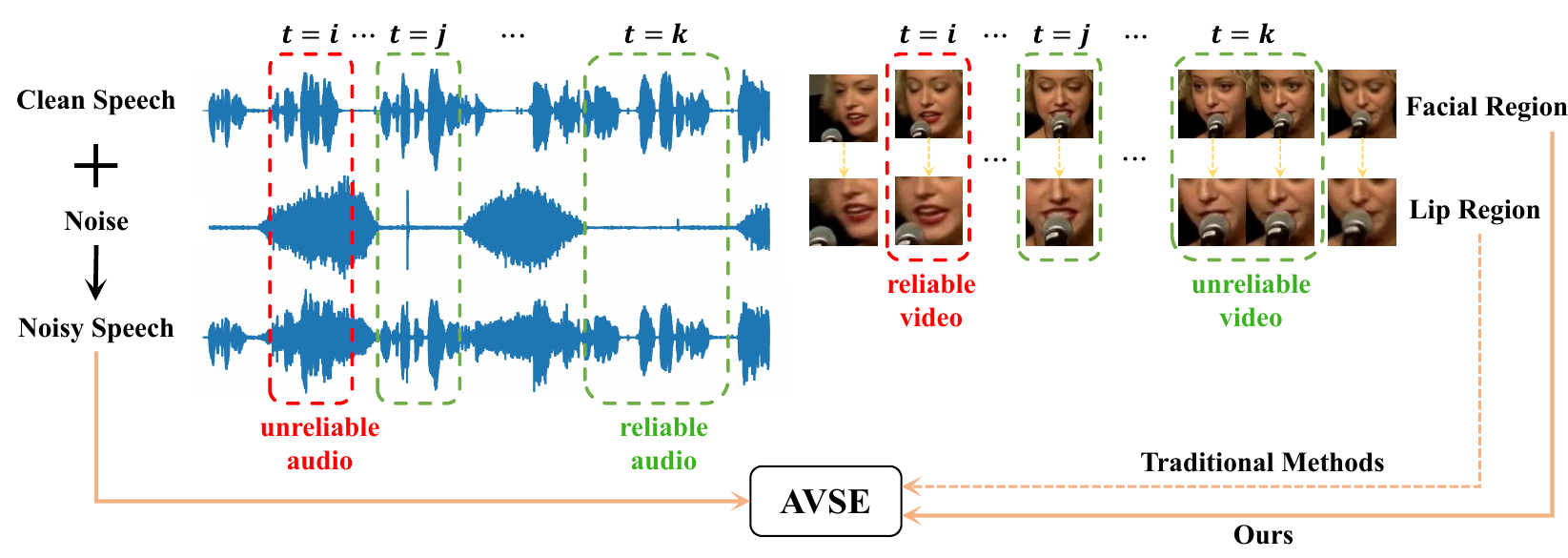}
    \vspace{-1.6em}
    \caption{\textbf{Illustration of our idea.} Beyond lip motion, 
    the facial region contains abundant speech-related information such as gender, age, nationality, and skin color, which can reflect the tone or accent of a speaker. Furthermore, such helpful information in the facial region is always hard to conceal compared with the small lip region. This leads to the main motivation for using facial cues in our method. However, there also exists speech-unrelated information in the facial region such as background, glasses, microphone, and speaker's hand movements.
    Additionally, some real-world issues, such as lip occlusion, head pose rotation, low resolution, and so on, would cause visual quality to vary during speech. At the same time, the non-stationary noise makes the audio quality also change significantly with time. These observations and analysis lead to our dual attention cooperative framework for AVSE.
    }
    \label{fig:overview}
\end{figure}\vspace{-1em}

Existing AVSE methods~\cite{hou2018audio, michelsanti2019deep, afouras2019my, iuzzolino2020av, wang2020robust} mostly take the speaker's lip region as visual input to capture semantic information for assisting speech enhancement. This additional information from visual modality remarkably improves the performance of speech enhancement.
However, extracting the accurate lip region is typically challenging due to the common occurrence of lip occlusion and low-resolution issues in practical scenarios.
%

Beyond the lip region, the facial region contains additional abundant information that is beneficial for speech enhancement, such as the speaker's gender, age, nationality, skin color, and so on, which can reflect the speaker's vocal tone, accent, and other characteristics related to speech.
Some works have preliminarily investigated the manners to use the face for speech enhancement tasks~\cite {chuang2020lite, gao2021visualvoice, afouras2019my}. However, the utilization of facial information for AVSE is still very limited and challenging.
%
As shown by \cite{chuang2020lite}, the full face image contains redundant irrelevant information for the speech enhancement task and if facial information is not utilized properly it will not improve the performance for AVSE.
%
%
Specifically, the speaker's facial images may contain decorative objects like glasses or backgrounds that are unrelated to the speech itself.
How to effectively utilize facial cues is an important and challenging problem for effective speech enhancement.

In this paper, to extract valuable information from facial regions, we \textbf{firstly} developed a visual feature extractor equipped with a spatial attention module.
This extractor aims to collect the global context from the whole facial region, instead of focusing only on a local area, and meanwhile ignoring the irrelevant redundant information.
This global context includes both the lip motion which is directly correlative to the audio speech information, and the facial appearance characteristics which implies the speech traits like tone, accent, and so on.
\textbf{Secondly}, it is widely acknowledged that during the speech, speakers tend to naturally gesture with their heads and facial expressions. This means that the degree to which visual cues assist with speech enhancement is constantly changing. Based on this observation, a dynamic visual feature fusion strategy is proposed to consider the reliability of visual features at different time steps and to reduce unreliable visual information during the modal fusion stage.
\textbf{Thirdly}, unstable noise in real-world scenarios always causes significant fluctuations in speech quality. Therefore, a dynamic audio feature fusion strategy is finally introduced to measure the dependability of audio across various temporal segments.

Based on these three points mentioned above, we propose our cooperative dual attention framework for AVSE, named DualAVSE. For the visual encoder, the dual attention mechanism is reflected in the process of both extracting robust visual features and measuring the reliability of visual features in the temporal dimension during the modal fusion stage. For the audio-visual fusion module, the dual attention mechanism is reflected by the dynamic fusion strategy in both the visual and audio modalities based on the temporal dimensional attention module. By integrating these two cooperative dual attention mechanisms, our method is robust for the speech-unrelated facial cues and shows advantages for the AVSE task under several different settings. 


In summary, our main contributions are as follows: \textbf{(1)} Unlike traditional methods that rely solely on the lip region, we explore leveraging facial cues for AVSE. We propose a novel cooperative dual attention framework to take full advantage of both facial and audio cues.
\textbf{(2)} We introduce the dual attention mechanism cooperated in two aspects, including the process of the visual encoder itself and the dynamic fusion of the audio-visual modalities.
\textbf{(3)} Our approach not only surpasses existing methods evaluated under multiple metrics but also demonstrates robustness to the challenge of unreliable or even absent input videos.
%

\sectionvspace
\section{Related Work}
\vspace{-1em}
\subsection{Audio-Only Speech Enhancement}
\subsectionbelowvspace
Audio-only speech enhancement aims to improve the quality and intelligibility of audio speech signals and plays an important role in various applications such as hearing aids~\cite{venema2006compression,levit2001noise,chern2017smartphone}, teleconferencing, speech recognition~\cite{li2015robust,li2013investigation}, and so on.
Most Traditional methods including Spectral Subtraction~\cite{boll1979suppression}, Wiener Filtering~\cite{lim1978all}, and Minimum Mean Squared Error~\cite{ephraim1984speech} are based on statistical assumptions, handling stationary noise well.
In recent years, deep learning-based methods have shown promising results for AOSE.
According to the manner to obtain enhanced speech, existing speech enhancement technologies can be divided into two categories: mask-based methods~\cite{wang2013towards, wang2014training, williamson2015complex} and mapping-based methods~\cite{lu2013speech, liu2014experiments, xu2014regression, kolbaek2016speech, fu2017complex, strake2020fully, fu2018end, pandey2019new, defossez2020real, thakker2022fast}.
~\cite{wang2013towards} estimates an ideal binary mask (IBM) to indicate the presence or absence of speech at each time-frequency bin, which has been one of the most classical methods and greatly promotes the development of speech enhancement afterward. A variant of IBM ideal ratio mask (IRM)~\cite{wang2014training} is proposed to indicate the desired signal-to-noise ratio (SNR) at each time-frequency bin. Another variant of IBM complex ideal ratio mask (cIRM)~\cite{williamson2015complex} operates on the complex domain instead of the magnitude domain, which represents both the amplitude and phase at each bin and improves the performance of AOSE.
%
Different from the mask-based methods, the mapping-based method directly estimates the enhanced speech and can be classified into spectrum mapping~\cite{lu2013speech, liu2014experiments, xu2014regression, kolbaek2016speech}, complex spectrum mapping~\cite{fu2017complex, strake2020fully}, waveform mapping~\cite{fu2018end, pandey2019new, defossez2020real, thakker2022fast}, etc. based on the input type.
These AOSE methods based on deep learning have significantly surpassed traditional enhancement algorithms due to their excellent non-linear mapping capabilities.
Additionally, it has achieved good denoising effects for non-stationary noise in real-world scenarios. Given the advantages of mask-based methods, in this paper, we employ cIRM for predicting enhanced audio.
\subsectionvspace
\subsection{Audio-Visual Speech Enhancement}
\subsectionbelowvspace
Inspired by McGruk~\cite{mcgurk1976hearing} and the Cocktail effect~\cite{cherry1953some}, researchers have begun attempting to introduce visual cues in the speaking process together with the audio signal for speech enhancement in recent years.
%
%
Since lip motion can intuitively reflect semantic information during speech. Most existing AVSE works~\cite{hou2018audio, michelsanti2019deep, afouras2019my, iuzzolino2020av, wang2020robust, gao2021visualvoice, pan7muse, zhu2023real, iuzzolino2020av} extracted visual information based on target speaker's lip regions of interest (ROIs).
However, it is always hard to precisely extract lip ROIs and lip motion information when faced with real-world issues such as lip occlusion, head pose rotation, low resolution, and so on.
~\cite{afouras2019my} proposed the utilization of an enrollment audio of the target speaker to supplement missing discriminative information when the visual encoder experiences lip occlusion. This improves the model's robustness to lip occlusion and achieves good results.
~\cite{gao2021visualvoice} introduces a single-face image of the target speaker to provide a prior for what sound qualities to listen for, as a replacement for the enrollment audio to mitigate this issue. 
In this work, we propose to extract global information from the speaker's facial video. In addition, considering the dynamic effect of the speaker's characteristics on the speech enhancement in real-world scenarios, we introduce a dynamic fusion strategy for visual features to compute reliability at different time steps, which is then utilized for guiding the subsequent audio-visual fusion. As the non-station of the noise and speech, we also employ the dynamic fusion strategy in the audio encoder.
\subsectionvspace
\subsection{Audio-Visual Speech Analysis}
\subsectionbelowvspace
Audio-visual speech analysis is a well-established field that aims to extract information from both the visual and audio modalities during speech production. Most research in this area has focused on audio-visual speech recognition (AVSR)~\cite{tamura2015audio, yu2020audio, afouras2018deep, noda2015audio, stewart2013robust}, where the goal is to recognize spoken words by combining information from both audio and visual cues, with emphasis on lip movements due to their importance in conveying phonetic information. The remarkable performance that these methods have achieved illustrates the significance of visual speech information. In addition to these lip-based AVSR, some researchers have explored the visual speech information beyond the lip~\cite{zhang2022lip, zhang2020can}. These researches demonstrate that leveraging face as input yields significantly better performance compared to traditional methods taking the lip ROIs. Their works also inspire us to utilize facial cues for AVSE.

\sectionvspace
\section{DualAVSE}
\begin{figure}[h]
    \vspace{-0.3em}
    \centering
    \includegraphics[width=12cm]{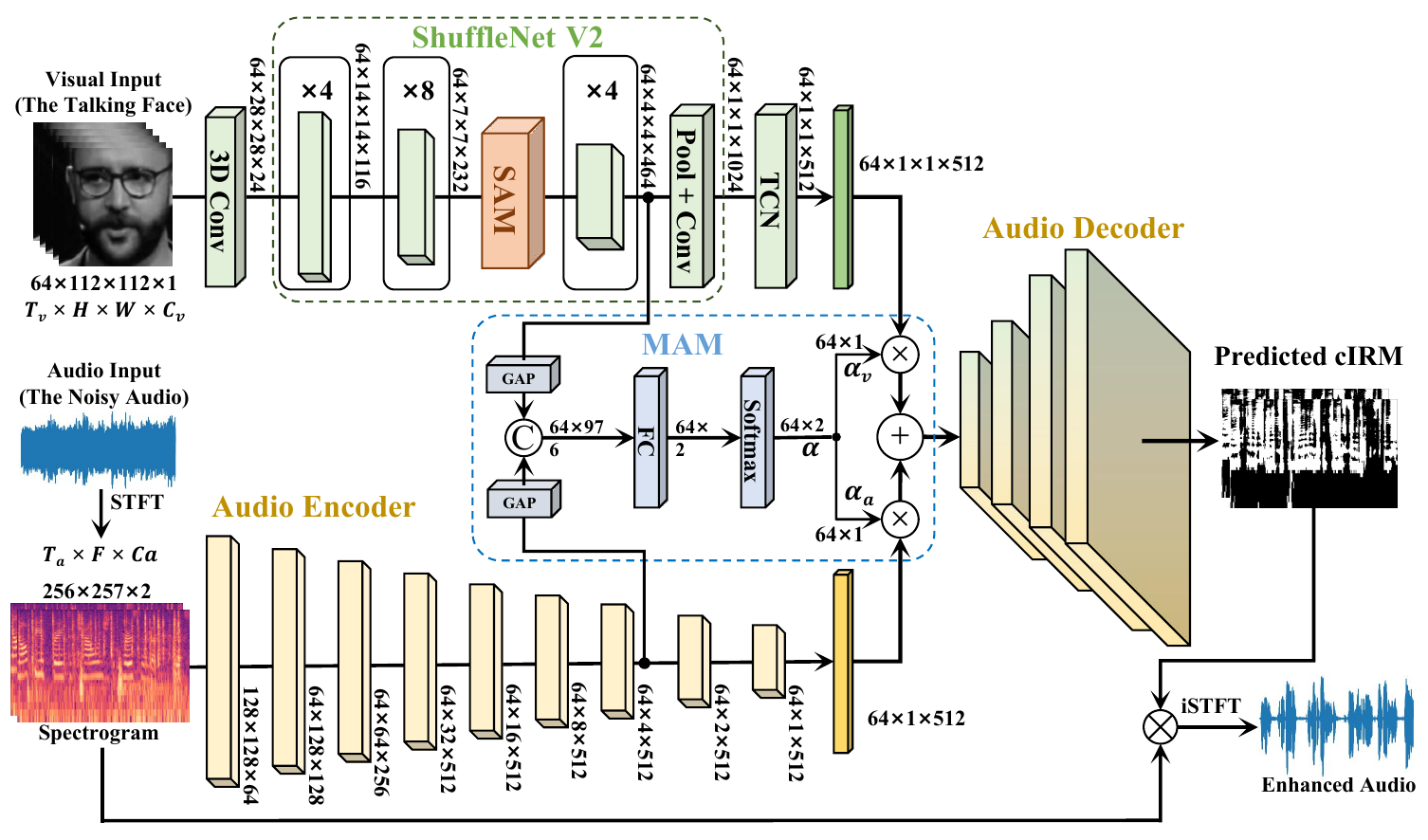}
    \vspace{-1em}
    \caption{Our proposed DualAVSE architecture.}
    \label{fig:model}
\end{figure}

\vspace{-0.8em}
In this section, we present DualAVSE for conducting speech enhancement with facial ROIs and noisy audio. Our DualAVSE framework consists of a visual encoder with a spatial attention module (SAM), a U-Net style audio codec, and a modality attention module (MAM), as indicated in Figure~\ref{fig:model}. 

\subsectionvspace
\subsection{Visual Encoder}\label{sec:visual_encoder}
\subsectionbelowvspace
Inspired by the visual speech recognition works~\cite{ma2021towards, martinez2020lipreading}, the backbone of our visual encoder contains a 3D convolutional layer that performs downsampling on the input image sequence in the spatial domain. Subsequently, we employ a lightweight ShuffleNet V2 network to accelerate the model convergence without compromising its performance. Finally, a Temporal Convolutional Network (TCN) is used to model the temporal dependencies on the output features of ShuffleNet V2. The visual features output by TCN has a dimension of $C_v\times T_v$, where $C_v$ is the channel dimension and $T_v$ is the time dimension.

\textbf{Spatial Attention Module.} To efficiently capture global contextual information from the entire facial image, extract potential speech-related features from the facial region, and avoid and potential interference from the speech-unrelated information, we introduce a spatial attention module (SAM) based on self-attention in the auxiliary network. We implement SAM with a single self-attention layer for simplicity here.
SAM scrutinizes every pixel of the input in the spatial domain, establishing connections with all other pixels within the same frame of the feature map. Thus SAM has the potential to capture the spatial context of the whole face, which enables the subsequent networks to relatively easily obtain beneficial information from regions beyond the lips. 

\textbf{Dynamic Fusion Strategy for the Visual Feature.} Considering the dynamic variations of the video quality across the time steps, we introduce a dynamic fusion strategy for integrating the visual encoder's output features. We combine the intermediate features from the visual encoder and the audio encoder (in Sec~\ref{sec:audio_encoder}) to generate an attention vector $alpha_v$, which aims to measure the reliability of visual and audio modalities.
The dimension of $\alpha_v$ is $T_v\times 1$. $\alpha_v$ is applied to assign a weight to each frame's visual feature before the final fusion of audio-visual features. Further details to fuse the visual features together with the audio features will be presented in section~\ref{sec:MAM}.

\renewcommand{\dblfloatpagefraction}{1.0}
\vspace{-1em}

\subsectionvspace
\subsection{Audio Encoder}\label{sec:audio_encoder}
\subsectionbelowvspace
For the audio encoder, it takes the complex spectrum $S_{noisy}$ obtained by applying the Short-Time Fourier Transform (STFT) to the noisy audio $s_{noisy}$ as input. $S_{noisy}$ has the dimensions of $2\times F\times T$, where $F$ and $T$ represent the frequency and time dimensions of the spectrum, respectively.
The encoder is composed of 9 convolutional layers and 7 average pooling layers as shown in Figure~\ref{fig:model}, which would downsample the input spectrum's frequency dimension to 1 and the time dimension to $T_a$. The final output feature has a dimension of $C_a\times T_a$, where $C_a$ and $T_a$ denote the channel and temporal dimension respectively.

\textbf{Dynamic Fusion Strategy for the Audio Feature.} In real-world scenarios, non-stationary noise exhibits diverse variations, leading to significant fluctuations in speech quality over time. We employ a similar structure as the visual encoder to obtain an attention vector $\alpha_a$, whose dimension is $T_a\times 1$. It is then applied to the audio feature of the audio encoder before fusing audio-visual features. Further details will be presented in section~\ref{sec:MAM}

\subsectionvspace
\subsection{Modality Attention Module}\label{sec:MAM}
\subsectionbelowvspace
As discussed in section~\ref{sec:visual_encoder} and ~\ref{sec:audio_encoder}, in real-world scenarios, the reliability of both audio and visual modalities varies significantly over time.
Based on this observation, we have introduced the dynamic fusion strategy to integrate the audio and visual features, leading to the design of our modality attention module (MAM).
The intermediate features from both the audio ($m_a$) and visual ($m_v$) encoders are reduced in dimensionality using Global Average Pooling (GAP). Afterward, they are fused by concatenation. The fused features are passed through a fully connected layer followed by a softmax activation function, resulting in a $2\times N$ attention vector $\alpha$, where $N$ represents the number of frames, $N=T_a=T_v$. \textbf{$\alpha$} is calculated as~\ref{eq:MAM}. We employ a learnable temperature parameter ($t$) to sharpen \textbf{$\alpha$} as below,
\vspace{-1em}
\begin{equation}\label{eq:MAM}
\boldsymbol{\alpha}={\rm Softmax}\left(\frac{{\rm FC}([{\rm GAP}(m_v);{\rm GAP}(m_a)])}{t}\right).
\end{equation}

\vspace{-0.8em}
During modality fusion, this attention vector was used to weigh the modality reliability at each time step as below:
\begin{equation}\label{eq:fusion}
f_{av}=f_v\otimes\alpha_v+f_a\otimes\alpha_a,
\end{equation}
where $f_{av}$ is the fused audio-visual feature, $\otimes$ denotes Element-wise multiplication at each time step. The fused features are then passed to the decoder to generate $M_p$.
\subsectionvspace
\subsection{Audio Decoder}
\subsectionbelowvspace
The audio decoder adopts a symmetric structure to the audio encoder. It takes the fused audio-visual features as input and goes through a series of upsampling operations to ultimately output a predicted cIRM $M_p$ with a dimension of $2\times F\times T$, the same as the input spectrum. Subsequently, the predicted $M_p$ is multiplied with the input spectrogram in the complex domain to obtain the predicted complex spectrogram. Finally, the enhanced audio is obtained by performing the inverse Short-Time Fourier Transform (iSTFT) on the predicted complex spectrogram.

\subsectionvspace
\subsection{Training Objective}
\subsectionbelowvspace
Our model predicts a complex mask $M_p$ to estimate the speech of the target speaker. Our training objective is to minimize the distance between the predicted mask $M_p$ and the ground-truth mask $M_{gt}$. $M_{gt}$ is calculated as below:
\vspace{-1em}
\begin{equation}\label{eq:mgt}
M_{gt}=S_{clean}*S_{noisy}^{-1}, 
\end{equation}
where $*$ denotes complex domain multiplication, $S_{clean}$ denotes the complex spectrum of target clean speech, and $S_{noisy}^{-1}$ denotes the inverse of the complex spectrum of the input noisy audio. We minimize the L2 based loss as below:
\begin{equation}\label{eq:loss}\vspace{-0.2em}
L=||M_{gt}-M_{p}||_2. \end{equation}

\sectionvspace
\section{Experiments}

\vspace{-0.6em}
\subsection{Experimental Seetings}
\subsectionbelowvspace
\subsubsection{Datasets}
\vspace{-0.6em}
\textbf{LRS3~\cite{afouras2018lrs3}:} This dataset contains 438 hours of talking videos from TED and TEDX clips downloaded from YouTube. We evaluate our method on the pretrain subset which contains 407 hours of video. We partitioned this subset into training, validation, and testing sets with a ratio of 8:1:1. Each frame in the video underwent face detection using the 3D FAN~\cite{yang2020fan}, which allowed us to extract 68 facial landmark points. Then, Procrustes analysis was applied to perform an affine transformation on the target face. This transformation was used to align the target face with the mean face. The input image size for the face region is $112\times 112$. To compare the model with the lip region as input, we also extracted lip ROIs from each frame using the same method, resulting in $88\times88$ pixel-sized lip regions of interest.
\\
\textbf{DNS4~\cite{dubey2022icassp}:} We follow~\cite{defossez2020real} to obtain the noise signal from the noise subset of the DNS dataset. The subset contains approximately 181 hours of noise audio collected from a wide variety of events. During training and evaluation, we utilized these samples as background noise to add noise to the clean speech and construct synthetic noisy audio inputs.
\\
\textbf{GRID~\cite{cooke2006audio}, CHiME~\cite{barker2015third}:}
We also evaluate on GRID with CHiME3 benchmark datasets to compare our model with the state-of-the-art AVSE methods: L2L~\cite{ephrat2018looking}, VSE~\cite{gabbay2017visual}, OVL~\cite{wang2020robust}. The GRID dataset consists of 33 speakers. For our experiments, we follow the general setting~\cite{balasubramanian2023estimation} to designate speakers s2 and s22 as the validation set, speakers s1 and s12 as the unseen unheard test set, and the remaining 29 speakers as the training set. We sample noise from CHiME to corrupt the clean speech. The noise in CHiME
is categorized into 4 types: Cafe, Street, Bus, and Pedestrian. The CHiME dataset is divided into training and testing sets with an 8:2 ratio as~\cite{wang2020robust}.

Following the prevailing practice in speech enhancement domain~\cite{ephrat_looking_2018, gao2021visualvoice, afouras2018deep}, we use synthetic noisy samples to train and evaluate our models. This is achieved by combining the waveforms of two separate clips, where one clip contains clean speech from the target speaker and the other clip contains interfering audio in the form of background noise.

\subsubsectionvspace
\subsubsection{Evaluation Metrics}
\vspace{-0.8em}
For evaluating our methods, we use standard speech enhancement metrics involving Signal-to-Distortion-Ratio (SDR)~\cite{raffel2014mir_eval}, Short-Time Objective Intelligibility (STOI)~\cite{taal2011algorithm} and Perceptual Evaluation of Speech Quality (PESQ)~\cite{rix2001perceptual}. (i) SDR: It is a commonly used metric for evaluating the quality of speech enhancement algorithms. It measures the ratio of signal strength to distortion between the processed speech signal and the original clean signal.  (ii) STOI: It measures the intelligibility of the signal (from 0 to 1), higher is better. (iii) PESQ: It rates the overall perception quality of the output signal (from 0.5 to 4.5), higher is better.

\subsubsectionvspace
\subsubsection{Implementation Details}
\vspace{-0.8em}
Our AV speech enhancement framework is implemented in PyTorch. For all experiments, we sub-sample the audio at 16kHz, and the input speech segment is fixed to 2.55s long as~\cite{gao2021visualvoice}. STFT is computed using a Hann window with a length of $400$, a hop size of $160$, and an FFT window size of $512$. The complex spectrogram is of dimension $2\times 257 \times 256$. The audio feature output by the audio encoder is of dimension $C_a\times N$, with $C_a=512, N=64$. The input to the visual encoder is the face ROIs of the size of $112\times 112$ from a sequence of $N = 64$ frames (2.55s). The visual feature output by the visual encoder is of dimension $C_v\times T_v$ with $C_v=512, T_v=64$. The entire network is trained using an Adam optimizer with weight decay of 1e-4, and batch size of 56. During training, we randomly sample a speech segment and a noise segment in the training set to synthesize training samples of noisy audio. 

Our DualAVSE utilizes a three-stage training approach to fully leverage the strengths of both MAM and SAM modules. In the first stage, the entire network is trained until convergence. In the second stage, the audio encoder and visual encoder up to SAM are frozen, and the remaining parts are trained until convergence. In the third stage, the entire network is unfrozen and trained until convergence.

To simulate real-world noise environments, we set four different signal-to-noise ratio (SNR) conditions: low SNR (-15dB), moderate SNR (-10dB, -5dB), and high SNR (0dB).

\subsectionvspace
\subsection{Results}
\subsectionbelowvspace
In this section, we give a detailed evaluation of the proposed method, including ablations, robustness analyses, and comparison to baselines. We first compare the performance of our models when they are conditioned on different input modality combinations; we then perform robustness tests in settings where visual modality is unreliable; we finally compare to the SOTA on the speech enhancement tasks. We also provide extra quantitative and qualitative results in the supplementary material.

\subsubsectionvspace
\subsubsection{Ablation Study}
To better understand the influence of different components in the proposed model on the overall performance, we conducted an ablation study on the input type and the attention modules. The results of the ablation experiments are presented in Table~\ref{tab:ablation}.
%

\textbf{AVSE Baseline vs. AOSE Baseline.} The AVSE baseline is obtained by inserting a Visual Encoder without SAM and MAM into the Audio Encoder. It can be observed from the first three lines in Table~\ref{tab:ablation} that incorporating visual modality brings significant improvement.

\textbf{MAM.} After further incorporating MAM into the AVSE Baseline when using the face as input, the model's performance shows obvious improvements across all metrics compared to the AVSE Baseline. This suggests that MAM, compared to simple concatenation fusion, is more effective in leveraging the information from both modalities.

\textbf{SAM.} The model's performance significantly improved over the baseline when introducing the SAM, indicating that the introduction of global context allows the model to more fully exploit the visual modality information.

\textbf{DualAVSE.} The final comparative results demonstrate our method significantly outperforms the baseline and yields the best performance. Moreover, comparing the results of the models using face and lip inputs, it can be seen that using the face as input leads to greater improvement. This demonstrates that the additional information contained in the facial region can be effectively explored by our method to improve the performance of AVSE.

\begin{table}[htpb]
\setlength{\tabcolsep}{3pt}
\begin{center}
\vspace{-0.2em}
\scalebox{0.8}{\begin{tabular}{lcccccccccccc}
\hline
\multirow{2}{*}{Model} & \multicolumn{3}{c}{-15dB} & \multicolumn{3}{c}{-10dB} & \multicolumn{3}{c}{-5dB} & \multicolumn{3}{c}{0dB} \\
\cmidrule(r){2-4} \cmidrule(r){5-7} \cmidrule(r){8-10} \cmidrule(r){11-13} 
 & SDR & PESQ & STOI & SDR & PESQ & STOI & SDR & PESQ & STOI & SDR & PESQ & STOI \\
\hline
AOSE Baseline & 3.38 & 1.333 & 0.639 & 6.41 & 1.502 & 0.735 & 8.76 & 1.718 & 0.812 & 11.09 & 1.999 & 0.868 \\
AVSE Baseline input lip & 3.34 & 1.356 & 0.665 & 6.47 & 1.536 & 0.751 & 8.91 & 1.765 & 0.819 & 11.41 & 2.071 & 0.873 \\
AVSE Baseline input face & 3.79 & 1.363 & 0.676 & 6.89 & 1.540 & 0.759 & 9.34 & 1.770 & 0.825 & 11.62 & 2.082 & 0.877 \\
+MAM & 3.83 & 1.364 & 0.676 & 6.94 & 1.555 & 0.761 & 9.35 & 1.790 & 0.828 & 11.70 & 2.103 & 0.879 \\
+SAM & 4.19 & 1.406 & 0.695 & 7.28 & 1.606 & 0.776 & 9.73 & 1.864 & 0.839 & 12.12 & 2.186 & 0.887 \\
DualAVSE input lip & 4.25 & 1.402 & 0.693 & 7.32 & 1.603 & 0.775 & 9.77 & 1.858 & 0.839 & 12.11 & 2.190 & 0.888 \\
DualAVSE input face & \textbf{4.45} & \textbf{1.435} & \textbf{0.700} & \textbf{7.54} & \textbf{1.643} & \textbf{0.780} & \textbf{9.96} & \textbf{1.909} & \textbf{0.843} & \textbf{12.32} & \textbf{2.241} & \textbf{0.889} \\
\hline
\end{tabular}}
\end{center}
\vspace{-2em}
\caption{Ablation study for our Audio-Visual Speech Enhancement method on LRS3 dataset.}
\label{tab:ablation}
\end{table}\vspace{-1.5em}

\subsubsectionvspace
\subsubsection{Robustness to the Unreliable Visual Modality}
To explore the difference between using the face and lip region as input, we conducted a series of comparisons. As shown in Table~\ref{tab:robust_results}, we separately trained AVSE models using the face and lip region as input. During testing, we applied different visual masks to evaluate the robustness of the model. For the face input, we used four approaches: \textbf{Fa} normal face input; \textbf{Fb} no face input; \textbf{Fc} random mask of the face video; and \textbf{Fd} occlusion of the lip area in the face input. For the lip input, we used three approaches: \textbf{La} normal lip input; \textbf{Lb.} no lip input; and \textbf{Lc} random mask of the lip video. For the random mask of the video, time and spatial dimensions are both randomly selected from 0 to 100\%.

The comparison of \textbf{Fb} and \textbf{Lb} shows that when the visual modality information is removed from the AVSE model in the testing process, the model trained with face input performs better. This intriguingly suggests that using the face as input can indeed more competently assist the model in learning the audio modality, thereby enabling it to extract more useful information for speech enhancement. 

Comparing the results of \textbf{Fd} and \textbf{Lb}, it can be seen that regions of the face other than the lip area can also effectively assist the model in speech enhancement. Here we calculate the performance gain by calculating the average of all metrics under all SNRs. All subsequent calculations follow the same methodology. Their performance degradation relative to their respective baselines is \textbf{1.62\%} and \textbf{2.43\%}, respectively, indicating that using the face as input has good robustness to lip occlusion issues. 

A similar conclusion can be drawn by comparing the results of \textbf{Fc} and \textbf{Lc}. Random masking of the face causes a performance degradation of \textbf{0.46\%}, lower than the degradation of a random mask of the lip area, which accounts for \textbf{0.81\%}. As under the same random masking ratio, masking the face typically covers a larger area than masking the lip region. These results further highlight the robustness of using face input.

\vspace{-0.8em}
\begin{table}[htpb]
\setlength{\tabcolsep}{3pt}
\begin{center}
\scalebox{0.8}{\begin{tabular}{lcccccccccccc}
\hline
\multirow{2}{*}{Model} & \multicolumn{3}{c}{-15dB} & \multicolumn{3}{c}{-10dB} & \multicolumn{3}{c}{-5dB} & \multicolumn{3}{c}{0dB} \\
\cmidrule(r){2-4} \cmidrule(r){5-7} \cmidrule(r){8-10} \cmidrule(r){11-13} 
 & SDR & PESQ & STOI & SDR & PESQ & STOI & SDR & PESQ & STOI & SDR & PESQ & STOI \\
\hline
\textbf{Fa:} Reliable Face & \textbf{4.45} & \textbf{1.435} & \textbf{0.700} & \textbf{7.54} & \textbf{1.643} & \textbf{0.780} & \textbf{9.96} & \textbf{1.909} & \textbf{0.843} & \textbf{12.32} & \textbf{2.241} & \textbf{0.889} \\
\textbf{Fb:} mask whole face & 3.88 & 1.410 & 0.667 & 7.26 & 1.620 & 0.765 & 9.71 & 1.880 & 0.836 & 12.09 & 2.220 & 0.886 \\
\textbf{Fc:} mask lip in face & 4.16 & 1.418 & 0.681 & 7.37 & 1.628 & 0.771 & 9.83 & 1.891 & 0.838 & 12.23 & 2.223 & 0.888 \\
\textbf{Fd:} random mask face & 4.38 & 1.429 & 0.694 & 7.50 & 1.638 & 0.778 & 9.91 & 1.903 & 0.842 & 12.27 & 2.236 & \textbf{0.889} \\
\hline
\textbf{La:} Reliable Lip & 4.25 & 1.402 & 0.693 & 7.32 & 1.603 & 0.775 & 9.77 & 1.858 & 0.839 & 12.11 & 2.190 & 0.888 \\
\textbf{Lb:} mask whole lip & 3.86 & 1.385 & 0.664 & 7.06 & 1.581 & 0.760 & 9.53 & 1.839 & 0.832 & 11.90 & 2.169 & 0.884 \\
\textbf{Lc:} random mask lip & 4.16 & 1.397 & 0.688 & 7.28 & 1.600 & 0.772 & 9.34 & 1.852 & 0.838 & 12.08 & 2.186 & 0.887 \\
\hline
\end{tabular}}
\end{center}
\vspace{-2em}
\caption{Robustness to the unreliable visual modality.}
\label{tab:robust_results}
\end{table}\vspace{-1.5em}

\subsubsectionvspace
\subsubsection{Comparison with Others}
\vspace{-0.5em}
Since we utilize the DNS dataset as the noise, we compare with the noise suppression techniques Sudo rm -rf~\cite{tzinis2022remixit} provided on the DNS benchmark. Table~\ref{tab:comparison_aose} shows that DualAVSE significantly outperforms Sudo rm -rf~\cite{tzinis2022remixit}.

\vspace{-0.8em}
\begin{table}[htpb]
\setlength{\tabcolsep}{3pt}
\begin{center}
\scalebox{1.0}{\begin{tabular}{lccc}
\hline
\multirow{2}{*}{Model} & \multicolumn{3}{c}{0dB} \\
\cmidrule(r){2-4}
 & SDR & PESQ & STOI \\
\hline
Sudo rm -rf~\cite{tzinis2022remixit} & 7.65 & 1.462 & 0.822 \\
DualAVSE & \textbf{12.32} & \textbf{2.241} & \textbf{0.889} \\
\hline
\end{tabular}}
\end{center}
\vspace{-2em}
\caption{Results on the LRS3 dataset with noise from DNS4.}
\label{tab:comparison_aose}
\end{table}\vspace{-1.3em}

Because there are not many methods available for AVSE and almost all the methods are trained and tested on different datasets without a unified testing set. We reproduce several state-of-the-art open-source methods to perform comparison as shown in Table~\ref{tab:comparison1}.

We reproduce VisualVoice~\cite{gao2021visualvoice}, MuSE~\cite{pan7muse}, DEMUCS~\cite{defossez2020real} and evaluate them on LRS3 + DNS4 datasets. We also adapt DEMUCS to AVSE by adding the same visual encoder as ours (3D front-end + ShuffleNet V2 + TCN) which encodes the video into temporal features that are then concatenated with the audio features from the original audio encoder. We refer to this model as AV-Demucs. All models were implemented based on official open-source code and trained until convergence according to the original paper.
We perform comparison to the previous methods in Table ~\ref{tab:comparison1}. They are all evaluated on LRS3 + DNS4 datasets. For all noise conditions, DualAVSE outperforms other approaches in quality and intelligibility, achieving significant improvements across all metrics.

We also perform a comparison with existing AVSE methods on  GRID + CHiME: L2L~\cite{ephrat2018looking}, VSE~\cite{gabbay2017visual} and OVA~\cite{wang2020robust}. As shown in Table~\ref{tab:comparison2}, our model yields superior performance across all noise conditions.

\vspace{-0.5em}
\begin{table}[h]
\setlength{\tabcolsep}{3pt}
\begin{center}
\scalebox{0.8}{\begin{tabular}{lcccccccccccc}
\hline
\multirow{2}{*}{Model} & \multicolumn{3}{c}{-15dB} & \multicolumn{3}{c}{-10dB} & \multicolumn{3}{c}{-5dB} & \multicolumn{3}{c}{0dB} \\
\cmidrule(r){2-4} \cmidrule(r){5-7} \cmidrule(r){8-10} \cmidrule(r){11-13} 
 & SDR & PESQ & STOI & SDR & PESQ & STOI  & SDR & PESQ & STOI & SDR & PESQ & STOI \\
\hline
DEMUCS~\cite{defossez2020real} & 2.33 & 1.210 & 0.561 & 5.84 & 1.297 & 0.682  & 9.10 & 1.443 & 0.777 & 11.85 & 1.631 & 0.839 \\
AV-DEMUCS~\cite{defossez2020real} & 3.03 & 1.213 & 0.611 & 6.15 & 1.314 & 0.694 & 9.47 & 1.483 & 0.787 & 11.86 & 1.666 & 0.843 \\
MuSE~\cite{pan7muse} & -1.02 & 1.160 & 0.568 & 2.82 & 1.230 & 0.648  & 5.97 & 1.320 & 0.731 & 8.53 & 1.460 & 0.797 \\
VisualVoice~\cite{gao2021visualvoice} & 2.52 & 1.317 & 0.643 & 5.73 & 1.475 & 0.735 & 8.16 & 1.682 & 0.808 & 10.32 & 1.963 & 0.865 \\
\hline
DualAVSE & \textbf{4.45} & \textbf{1.435} & \textbf{0.700} & \textbf{7.54} & \textbf{1.643} & \textbf{0.780} & \textbf{9.96} & \textbf{1.909} & \textbf{0.843} & \textbf{12.32} & \textbf{2.241} & \textbf{0.889} \\
\hline
\end{tabular}}
\end{center}
\vspace{-2em}
\caption{Comparison on the LRS3 dataset with noise from DNS4.}
\label{tab:comparison1}
\end{table}

\vspace{-2em}
\begin{table}[htbp]
\begin{center}
\begin{tabular}{lccccccc}
\hline
SNR & -5dB & 0dB & 5dB & 10dB & 15dB & 20dB & Avg \\
\hline
L2L~\cite{ephrat2018looking} & 2.02 & 2.58 & 2.92 & 3.16 & 3.32 & 3.50 & 2.92 \\
VSE~\cite{gabbay2017visual} & 2.04 & 2.54 & 2.81 & 3.00 & 3.12 & 3.22 & 2.79 \\
OVA~\cite{wang2020robust} & 1.99 & 2.59 & 2.98 & 3.28 & 3.51 & 3.67 & 3.00 \\
DualAVSE & \textbf{2.16} & \textbf{2.67} & \textbf{3.06} & \textbf{3.43} & \textbf{3.79} & \textbf{4.05} & \textbf{3.19} \\
\hline
\end{tabular}
\end{center}
\vspace{-2em}
\caption{PESQ results On GRID dataset with noise from CHiME. Higher is better.}
\label{tab:comparison2}
\end{table}\vspace{-1.3em}


Furthermore, we conduct comparisons with the methods AV c-ref~\cite{morrone2019face} and VS~\cite{afouras2019my}, which also leverage facial cues for audio-visual speech separation (AVSS). For a fair comparison, we also input our model with two speaker inputs for training and testing. The results in Table~\ref{tab:face} and Table~\ref{tab:my} clearly demonstrate that DualAVSE outperforms both~\cite{morrone2019face} and~\cite{afouras2019my}.

\vspace{-0.4em}
\begin{table}[htbp]
\setlength{\tabcolsep}{5pt}
\begin{floatrow}
\capbtabbox{
\begin{tabular}{ccc}
\hline
Models & SDR & PESQ \\
\hline
AV c-ref~\cite{morrone2019face} & 8.05 & 2.70 \\
DualAVSE & \textbf{9.24} & \textbf{2.75} \\
\hline
\end{tabular}
}{
    \vspace{-3mm}
    \caption{Comparison with AV \\c-ref~\cite{morrone2019face} on GRID.}
    \label{tab:face}
}
\capbtabbox{
\begin{tabular}{cc}
\hline
Models & SDR  \\
\hline
VS~\cite{afouras2019my} & 12.8 \\
DualAVSE & \textbf{13.4} \\
\hline
\end{tabular}
}{
    \vspace{-3mm}
    \caption{Comparison with VS~\cite{afouras2019my} on LRS3.}
    \label{tab:my}
}
\end{floatrow}
\end{table}\vspace{-1em}

\vspace{-1em}
\section{Ethical Discussion}
\vspace{-0.8em}
Despite numerous positive applications, our method can also be misused. For example, audio-visual speech enhancement techniques can be used for eavesdropping. 

For this academic research, we utilize only publicly available datasets. We aim to approach this work ethically within the constraints of an academic research environment, in hopes of responsibly advancing speech enhancement. 

\vspace{-1.5em}
\section{Conclusion}
\vspace{-0.8em}
In this paper, we presented a robust Audio-Visual Speech Enhancement (AVSE) framework that leverages facial cues beyond the lip region. By incorporating global facial context and dynamic fusion strategies for visual and audio features with dual attention mechanisms, our model effectively captures speech-related information and mitigates the impact of noise and irrelevant attributes. The experimental results demonstrate the superior performance of our approach under various noise conditions and challenging scenarios. Our work showcases the robustness and effectiveness of utilizing facial cues in AVSE tasks, paving the way for improved speech enhancement systems in real-world settings.

\vspace{-1.3em}
\section{Acknowledgements}
\vspace{-0.8em}
This work is partially supported by National Natural Science Foundation of China (No. 62276247, 62076250).

\newpage
\bibliography{egbib}
\end{document}